\documentclass[a4paper,conference]{IEEEtran}

\usepackage{times}

\usepackage{epsfig}
\usepackage{graphicx}

\usepackage{amsmath}
\usepackage{amssymb}
\usepackage{bm}
\usepackage{tabularx}
\usepackage{float}

\usepackage{algorithm, algpseudocode}

\usepackage{hhline}
\usepackage{multirow}

\renewcommand{\figurename}{Fig.}
\renewcommand{\tablename}{Table }

\algrenewcommand\algorithmicrequire{\textbf{INPUT:}}
\algrenewcommand\algorithmicensure{\textbf{OUTPUT:}}

\begin{document}

\title{Spatially-weighted Anomaly Detection}

\author{
\IEEEauthorblockN{Minori Narita$^{1}$, \:Daiki Kimura$^{2}$, \:Ryuki Tachibana$^{2}$}
	\IEEEauthorblockA{$^{1}$The University of Tokyo, $^{2}$IBM Research AI\\
		Email: narita@g.ecc.u-tokyo.ac.jp, \{daiki, ryuki\}@jp.ibm.com}
}

\maketitle
\thispagestyle{empty}

\begin{abstract}

Many types of anomaly detection methods have been proposed recently, and applied to a wide variety of fields including medical screening and production quality checking. Some methods have utilized images, and, in some cases, a part of the anomaly images is known beforehand. However, this kind of information is dismissed by previous methods, because the methods can only utilize a normal pattern. Moreover, the previous methods suffer a decrease in accuracy due to negative effects from surrounding noises. In this study, we propose a spatially-weighted anomaly detection method~(SPADE) that utilizes all of the known patterns and lessens the vulnerability to ambient noises by applying Grad-CAM, which is the visualization method of a CNN. We evaluated our method quantitatively using two datasets, the MNIST dataset with noise and a dataset based on a brief screening test for dementia. 

\end{abstract}

\section{Introduction}

\subsection{Anomaly detection by images}

\begin{figure}[tb]
	\begin{center}       
		\includegraphics[width=80mm]{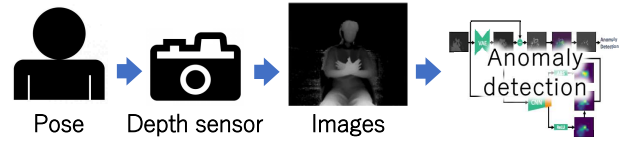}
	\end{center}
	\caption{Application of anomaly detection to YFPIT}
	\label{fig:overview}
\end{figure}

These days, many types of anomaly detection methods have been proposed and applied to fields such as production quality checking and medical screening~\cite{ADSurvey, MedicalAD, creditcard}. In real situations, a part of the anomaly patterns is often known beforehand. For example, various typical anomaly patterns from participants of a screening test for dementia called the Yamaguchi Fox-Pigeon Imitation Test (YFPIT)~\cite{yamaguchi} have been reported. With this test, participants made specific gestures to imitate a fox or a pigeon, and they were then screened on the basis of the result. Fig.~\ref{fig:overview} gives the overview of a method where anomaly detection is applied to YFPIT.

A straightforward approach is to train a model for only normal patterns, and then to classify by deviation. When using images for detection, it is common to train an auto-encoder such that reconstruction errors toward the normal images are minimized, and then use the error for detection. An advantage of this method is that users can check the results of the detection with their own eyes. However, they cannot utilize the information of anomaly patterns, even if they already know it. Moreover, the results might fluctuate due to error from noises, the model will make the noise values average.

Another approach is to solve the problem by training a classification model for both the normal and anomaly patterns. If the input is images, it is natural to perform by using a convolutional neural network~(CNN). The advantage of this approach is that all of the known patterns can be utilized. On the other hand, there is a serious vulnerability in that the output of the anomaly patterns is intrinsically unpredictable. It is important to maintain robustness against unknown patterns, especially when we aim for application to medical fields. For these reasons, a new method is required that can mitigate ambient noises and maintain robustness against unknown patterns, while being based on anomaly detection.

\subsection{Spatial information in images}

To maintain robustness against unknown patterns, only a normal pattern should be used during the training process for anomaly detection. On the other hand, a method such as weighting a region that is important for detection by using all of the training data can be implemented to buffer the effect of surrounding noises and utilize the known patterns. Generally speaking, object recognition methods such as Faster R-CNN~\cite{FasterRCNN} and SSD~\cite{SSD} seem to be suitable in this context, but these methods need region information for training. This is problematic because labeling all of the objects by hand is daunting work, and also, it is difficult to check quantitatively if the best labeling has been done.

Recently, Class Activation Mapping~(CAM), which visualizes the region of interest~(ROI) from the gradients of a CNN~\cite{cam}, and Grad-CAM~\cite{gradcam}, which is a generalization of CAM, have been proposed. Grad-CAM obtains the ROI from the gradients flowing into the last convolutional layer of a CNN. The significant advantage of Grad-CAM is that the regional information can be obtained without having to train the data that has the regional information. 

\subsection{Proposed method}

\begin{figure}[tb]
	\begin{center}
		\includegraphics[width=85mm]{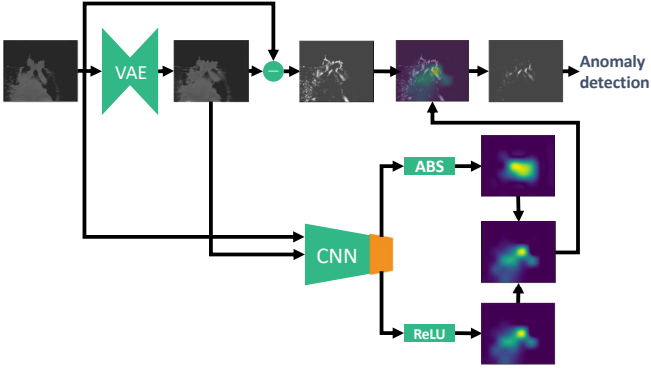}
	\end{center}
	\caption{Overview of SPADE}
	\label{fig:spade}
\end{figure}

In this paper, we propose a new method called SPatially-weighted Anomaly DEtection~(SPADE), which conducts anomaly detection by using Grad-CAM to weight the important region in an image. We also modified the Grad-CAM algorithm under this work. The overview of SPADE is presented in Fig.~\ref{fig:spade}.

\section{SPADE}

We assume that a part of the anomaly patterns is already known. Therefore, there are three classes in this problem setting: a normal class~$\mathbb{X}$, known anomaly classes~$\mathbb{A}$, and test data including unknown anomaly classes~$\mathbb{U}$. During the training process, $\mathbb{X}$ and $\mathbb{A}$ are used. $\mathbb{U}$ is used for evaluation.

The outline of the overall process is as follows. First, in the training process, a CNN is trained as the binary classifier and obtains the parameters related to the interest region information, and a Variational Auto-Encoder~(VAE) ~\cite{vae} is also trained using only the normal class. After that, the error weighted from the interest region in the image is obtained by multiplying the ROI given by the CNN using Grad-CAM and the reconstruction error given by VAE. Anomaly detection is performed depending on the value of this weighted error.

\subsection{Training}

The normal class~$\mathbb{X}$ is given to VAE as input, and parameters~$\bm{\theta}, \bm{\phi}$ of VAE are optimized to minimize the cost function

\begin{equation}
\begin{split}
\mathcal{L} (\bm{\theta}, \bm{\phi}, \bm{x_i} ) = &- D_{KL} \left( q _ { \bm{\phi} } ( \bm { z } | \bm{x_i} ) | | p _ { \bm{\theta} } ( \bm { z } ) \right) \\
&+ \mathbb { E } _ { q  ( \bm{ z } | \bm { x_i } ) } \left[ \log p _ { \bm{\theta} } ( \bm{ x_i } | \bm{ z } ) \right],
\end{split}
\end{equation}

where $\bm{x_i}$ is a sample data, $\bm{z}$ is a hidden space, $D_{KL}$ is KL divergence, and $\mathbb { E } _ { q  ( \bm{ z } | \bm { x_i } ) }\left[ \log p _ { \bm{\theta} } ( \bm{ x_i } | \bm{ z } ) \right]$ is the expected value of a logarithmic likelihood $\log p _ { \bm{\theta} } ( \bm{ x_i } | \bm{ z } )$ of input $x_i$ for an encoder $q$. The reconstruction error from input $\bm{e_x}$ is computed by these trained parameters $\bm{\theta}$ and $\bm{\phi}$.

At the same time, all of the training data~($\mathbb{X}+\mathbb{A}$), which are known patterns, are input to the CNN and trained to classify  a correct class and incorrect classes by optimizing the parameter~$\bm{\omega}$ to minimize the cost function below.

\begin{equation}
\mathcal{L} (\bm{\omega}, \bm{x_i}, t_i) = \| t_i - f(\bm{x_i}; \bm{\omega})\|,
\end{equation}

where $t_i$ is the lable of a sample $\bm{x_i}$, which has the value of 1 if $\bm{x_i}$ is the correct pattern, and 0 if not.

\subsection{Detection}

The detection algorithm is performed as follows. First, we calculate a reconstruction error $\bm{e_u}~(=| \bm{u}-\bm{\hat{u}}|)$ from a reconstructed image $\bm{\hat{u}}$ by inputting a test data $\bm{u}$ to the trained VAE network. At the same time, this test data $\bm{u}$ is also input to the CNN network and the ROI is estimated by using the visualization method, Grad-CAM. Grad-CAM is executed as follows. We first compute the gradient of the score for class $c$~(the normal pattern), or the partial derivative of the feature maps $A^k$ of a convolutional layer concerning $y^c$~(before the softmax layer)~(i.e., $\frac{\partial y^c}{\partial A^k}$). The gradient flowing back is global-average-pooled to obtain the importance weights of the $k_{th}$ feature map for the class $c$, $\alpha_k^c$.

\begin{equation}
 \alpha_k^c = \frac{1}{Z} \sum_i \sum_j \frac{\partial y^c}{\partial A^k_{ij}}
\end{equation}

Next, we perform a weighted combination of forward activation maps, and obtain the ROI, $\bm{L}_{CAM}^u$.

\begin{equation}
\bm{L}_{CAM}^u = f \left( \sum_k \alpha_k^c A^k \right),
\end{equation}

where $f$ is the activation function. $f$ is the absolute function (Abs) for input images, and the ReLU function for reconstruction images (described later). The eventual ROI $\bm{L}_{CAM}$ is

\begin{equation}
\bm{L}_{CAM} = \rm{Abs} \left( \sum_k \alpha_k^c A_u^k \right) + 
	\rm{ReLU} \left( \sum_k \alpha_k^c A_{\hat{u}}^k \right)
\end{equation}

We obtain a weighted reconstruction error $\bm{e'_u}$ focusing on the important region by multiplying $\bm{e_u}$ by $\bm{L}_{CAM}$. The input such that sum$(\bm{e'_u})$ is less than a threshold is classified as the correct pattern, and larger than the threshold is classified as an incorrect pattern. The outline of the algorithm above is shown in Algorithm~1. We describe the reasons and the details of the modification from a simple Grad-CAM below.

First, we took the sum of the ROI of a reconstructed image from a VAE as well as an original image. If we estimate the regional information only from the original image, we may fail to recognize the target object and underestimate the reconstruction loss for an anomaly pattern. Since a reconstruction image is close to a correct pattern, this modification enables us to weight the region that should be focused on.

\begin{figure}[tb]
	\begin{center}
		\includegraphics[width=5.0cm]{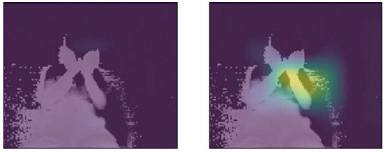}
	\end{center}
	\caption{Example of the difference of ROI between activation functions. The gray scale input image and the regional information image in RGB are overplotted. Left: Example activated by ReLU. Right: Example activated by Abs function.}
    \label{fig:gradcam-activation}
\end{figure}

Second, we changed the activation function of Grad-CAM from the ReLU function to the absolute function. Unlike Grad-CAM, in SPADE, we should focus on the region that has dissimilarity to the target as well. By activating by the absolute function, we can obtain the regional information that is significant for the detection more accurately. An example that fails to focus on the hand region when using the ReLU function is given in Fig.~\ref{fig:gradcam-activation}. Meanwhile, the activation function is ReLU when a reconstruction image is input. This is because the feature of a reconstructed image should always be close to the correct pattern.

\begin{figure}[tb]
	\begin{center}
		\includegraphics[width=5.0cm]{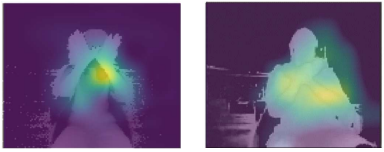}
	\end{center}
    \label{fig:norm}
	\caption{Difference of regional information（Left： Example which has small ROI，Right： Example which has large ROI）}
\end{figure}

Third, we normalized the weighting size. When the value of regional information (weighting) is large, the error will be large as well. By normalizing each ROI, we can get rid of the bias that comes from the region size. Examples of the difference between activation functions are shown in Fig.~\ref{fig:norm}.

We define SPADE as a method that weights the reconstruction error of VAE with the above modifications. The substantial difference from previous methods is that we utilize the regional information in images automatically by applying the visualization method of a CNN. This enables us to create a model robust against ambient noise unrelated to detection.

\algrenewcommand\algorithmicindent{0.8em}%
\begin{algorithm}[tb]
\caption{Spatially-weighted Anomaly Detection (SPADE)}
\label{algorithm3}
\begin{algorithmic}
\Procedure{Training}{}
\State $\bm{\phi, \theta, \omega} \gets$ \rm{Initialize parameters}

\Repeat
\State $\mathbb{X}^M \gets $ Random minibatch of M datapoints 
\State  $\forall \bm{x_i} \in \mathbb{X}^M,\ \ \bm{g} \gets \nabla_{\bm{\theta}, \bm{\phi}} \: {\mathcal{L}}(\bm{\theta}, \bm{\phi}, \bm{x_i})$  
\State $\bm{\phi, \theta} \gets$ Update parameters using gradients $\bm{g}$
\Until{convergence of parameters $(\bm{\phi}, \bm{\theta})$}

\Repeat
\State $\mathbb{(X+A)}^M \gets $ Random minibatch of M datapoints
\State  $\forall \{\bm{x_i}, t_i\} \in (\mathbb{X}+\mathbb{A})^M,\bm{g} \gets \nabla_{\bm{\omega}} \: {\mathcal{L}}(\bm{\omega}, \bm{x_i}, t_i)$ 
\State $\bm{\omega} \gets$ Update parameters using gradients $\bm{g}$
\Until{convergence of parameter $\bm{\omega}$}
\EndProcedure
\State 

\Procedure{Test}{}
\ForAll{$\bm{u} \in \mathbb{U}$}
\State $ \bm{\hat{u}} \gets R(\bm{u}; \bm{\theta, \phi})$ \ \ \ \ (obtain the reconstructed image)
\State $ \bm{e_u} \gets | \bm{\hat{u}} - \bm{u} | $
\State $ \bm{\bm{L}_{CAM}} \gets \text{Abs} \left( \sum_{k} \alpha^{c}_{\bm{u}} A^{k} \right) + \text{ReLU} \left( \sum_{k} \alpha^{c}_{\bm{\hat{u}}} A^{k} \right)$
\State $\bm{e'_u} \gets \bm{e_u} * \bm{L}_{CAM}\  / \ | \bm{L}_{CAM} |$
\If{sum$(\bm{e'_u}) > \:$thereshold}
\State Detect as incorrect pattern
\Else
\State Detect as correct pattern
\EndIf
\EndFor
\EndProcedure
\end{algorithmic}
\end{algorithm}

\section{Experiments}

\subsection{Comparative approach}

We evaluated the proposed method~(SPADE) and compared it with anomaly detection by the reconstruction error of a VAE~(VAE-based)~\cite{VAE-basedAD}, detection depending on the likelihood of a CNN~(CNN-based), and Na\"ive SPADE, which simply multiplies the reconstruction error of a VAE and the output of Grad-CAM. For Na\"ive SPADE, we computed regional information only from an input image, which is not normalized, and activated by the ReLU function.
In the experiment, we used the known normal and anomaly patterns for training data, and calculated the AUC of the ROC curve using all of the evaluation images including unknown anomaly patterns.

\subsection{MNIST with noise dataset}

\begin{figure}[tb]
	\begin{center}
		\includegraphics[width=5cm]{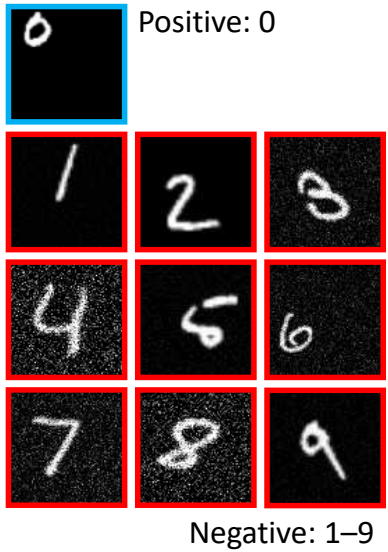}
	\end{center}
	\caption{MNIST with noises}
	\label{fig:mnist}
\end{figure}

In its original form, the MNIST dataset does not have noises, and a figure is written using the entire image space. Therefore, it is not suitable for this problem~(noises are included, and a target object exists in a specific position).
Hence, we added normalized noise to images and tripled the image size, and also changed the size and position of a figure depending on each image to create an MNIST dataset that includes noises.
Examples of images are shown in Fig.~\ref{fig:mnist}.
The value range of the MNIST dataset is $0$--$256$. We added the normalized noise by $\mathcal{N}(0,  \sigma^2)$, and computed the standard deviation $\sigma$ from $\mathcal{N}(40, 30^2)$. Image size was changed from $(28,28)$ to $(84,84)$, and the size and position were randomly chosen from inside this range.

As for the network structure, a VAE has $4$ convolutional layers and a $128$-dimensional hidden space, and a CNN consists of $3$ convolutional layers.
We assume that $0$ is the correct pattern, and $1$--$9$ are the incorrect patterns, and one of them is the known pattern while the others are the unknown patterns. Specifically, we computed the result in which we assume that one of $1, 3, 5, 7, 9$ is the known incorrect pattern.

The results are given in Table.~\ref{tab:result}. Our method has the comparative result than CNN-based baseline method.

\begin{table}[tb]
	\center
	\caption{AUROC for MNIST dataset with noises}
	\label{tab:result}
	\begin{tabular}{c|ccccc|c}
		\hline
		& 1 & 3 & 5 & 7  & 9 & Avg.\\
		\hline
		VAE-based\cite{VAE-basedAD} & .63 & .63 & .63 & .63 & .63 & .63 \\
		CNN-based & .73 & \bf{.88} & \bf{.96} & \bf{.88} & \bf{.96} & \bf{.88} \\
		Na\"ive SPADE & .67 & .59 & .65 & .76 & .53 & .64 \\
		\bf{SPADE} & \bf{.85} & .87 & .92 & .86 & .90 & \bf{.88} \\
		\hline
	\end{tabular}
\end{table}

\subsection{Hand gesture}

\begin{figure}[tb]
	\begin{center}
		\includegraphics[width=8cm]{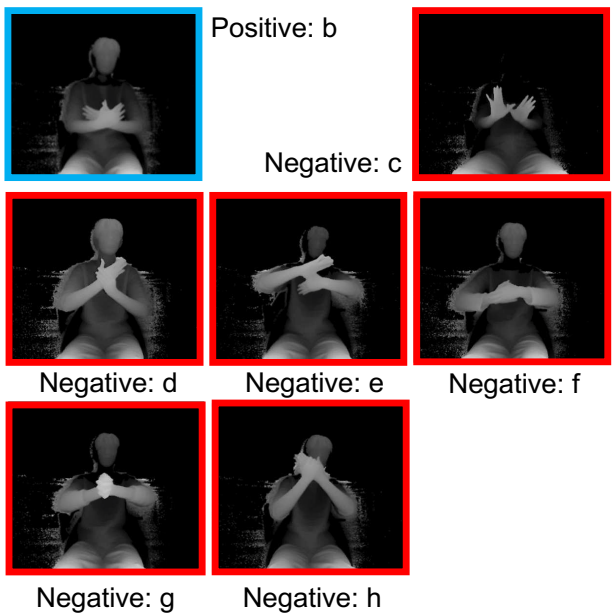}
	\end{center}
		\caption{Pigeon hand gesture dataset} 
		\label{fig:pigeon}
\end{figure}

We made a pigeon gesture dataset in reference to the Yamaguchi Fox-Pigeon Imitation Test~\cite{yamaguchi}.
The hand gesture dataset was created by taking $189,000$ images as a total, comprising seven types of gestures reported by \cite{yamaguchi} from $18$ participants who had different body shapes and gender. We used depth images from Kinect, as the shape and depth are more important  for detection than the color. An example of images is shown in Fig.~\ref{fig:pigeon}.

We took pictures following the process below. A participant sits on a chair and makes a gesture of each pattern while facing toward Kinect, and Kinect takes pictures continuously at the frequency of around $30$~fps. We had participants make the gesture while changing their position, angle of hands, and angle of posture. In addition, we asked participants to change their seated position on the chair, left, middle, and right. By collecting data comprising different types of hands and postures, we aimed to create a strongly robust classifier.

As for the network structure, the VAE has $4$~convolutional layers and a $256$-dimensional hidden space. We used the structure of ResNet\cite{resnet} for the CNN.
We assume that $b$~(pigeon gesture) is the correct pattern, and $c$--$h$ are the incorrect patterns, with one of them known and the others unknown. Specifically, we computed the result in which we assume that one of $c, d, e$ is the known incorrect pattern.

The results are given in Table.\ref{tab:p_result}. Our method outperformed the comparative methods for all of the given patterns.

\begin{table}[tb]
	\center
	\caption{AUROC for pigeon hand gesture dataset}
	\label{tab:p_result}
	\begin{tabularx}{80mm}{X|p{8.0mm}p{8.0mm}p{8.0mm}|p{8.0mm}}
		\hline
		& \hfil{c} & \hfil{d} & \hfil{e} & \hfil{Avg.}\\
		\hline
		\hfil{VAE-based\cite{VAE-basedAD}} & \hfil{.95} & \hfil{.95} & \hfil{.95} & \hfil{.95}\\
		\hfil{CNN\cite{resnet}-based} & \hfil{.86} & \bf{\hfil{.97}} & \hfil{.90} & \hfil{.91}\\
		\hfil{Na\"ive SPADE} & \hfil{.80} & \hfil{.71} & \hfil{.78} & \hfil{.76} \\
		\bf{\hfil{SPADE}} & \bf{\hfil{.98}} & \bf{\hfil{.97}} & \bf{\hfil{.96}} & \bf{\hfil{.97}}\\
		\hline
	\end{tabularx}
\end{table}

\section{Conclusion}

In this paper, we proposed a new method that utilizes the information of known anomaly patterns for an anomaly detection problem in which a part of the anomaly patterns is already known.
We verified that the proposed method which combines anomaly detection based on a VAE with Grad-CAM, which is the visualization method of a CNN, outperformed previous methods. The three points below are the main contributions of this paper.

\begin{itemize}
	\item We proposed an new anomaly detection method that utilizes information of the known anomaly patterns for a problem in which a part of the anomaly patterns is already known
	\item The proposed method outperformed previous methods including an anomaly detection method using a VAE and the method using the likelihood of a CNN
	\item We demonstrated through the results of real environment dataset that this method can be applied to a wide variety of real situations.
\end{itemize}

From these points, we conclude that our method has potential applications to many fields including production quality checking and fraud detection.

\end{document}